\documentclass[runningheads,a4paper]{llncs}
\usepackage{graphicx,color}
\usepackage{times}
\usepackage{epsfig}
\usepackage{amsmath,amssymb}
\usepackage{algorithm,algpseudocode}
\usepackage{cite}
\usepackage{comment}
\usepackage{xspace}

\usepackage{url}

\newcommand{\keywords}[1]{\par\addvspace\baselineskip
\noindent\keywordname\enspace\ignorespaces#1}
\newcommand{\argmin}{\operatornamewithlimits{argmin}}

\def\groundtruth{ground truth}
\def\coregistration{coregistration}
\def\cosegmentation{cosegmentation}

\newcommand{\refsec}[1]{Section~\ref{#1}}
\newcommand{\reffig}[1]{Figure~\ref{#1}}
\newcommand{\refeq}[1]{Equation~\ref{#1}}

\newcommand{\mycomment}[1]{}
\def\segs{\mathcal{S}}
\def\imgs{\mathcal{I}}
\def\defs{\mathcal{T}}

\newcommand\blfootnote[1]{%
  \begingroup
  \renewcommand\thefootnote{}\footnote{#1}%
  \addtocounter{footnote}{-1}%
  \endgroup
}

\begin{document}

\mainmatter  

\def\papertitle{Prior-based Coregistration and Cosegmentation} 
\title{\papertitle}

\titlerunning{\papertitle}

%
%

\author{$^{*}$Mahsa Shakeri$^{2,4}$, $^{*}$Enzo Ferrante$^{1}$, Stavros Tsogkas$^{1}$, Sarah Lippe$^{3,4}$, \\Samuel Kadoury$^{2,4}$, Iasonas Kokkinos$^{1}$, Nikos Paragios$^{1}$}

\authorrunning{Shakeri et al.}


\institute{$^1$CVN, CentraleSupelec-Inria, Universite Paris-Saclay, France, \\$^2$ Polytechnique Montreal, Canada \\$^3$University of Montreal, Canada,  \\$^4$ CHU Sainte-Justine Research Center, Montreal, Canada}

%
%

\toctitle{\papertitle}
\maketitle
\begin{abstract}
We propose a modular and scalable framework for dense \coregistration\ and \cosegmentation{} with two key characteristics: 
first, we substitute \groundtruth\ data with the semantic map output of a classifier;
second, we combine this output with population deformable registration to improve both alignment and segmentation. 
Our approach deforms all volumes towards consensus, taking into account image similarities and label consistency. 
Our pipeline can incorporate any classifier and similarity metric.
Results on two datasets, containing annotations of challenging brain structures, demonstrate the potential of our method.\blfootnote{$^*$Authors contributed equally}
\keywords{\coregistration{}, \cosegmentation{}, discrete optimization, priors.}
\end{abstract}

\section{Introduction}
In recent years, multi-atlas segmentation (MAS) has become a widely used image segmentation technique for biomedical applications~\cite{Iglesias2015Multi}. It uses an annotated dataset of atlases (images with their corresponding \groundtruth\ labels) to segment a target image. The atlases are first registered to the target; then the deformed segmentation masks are fused, generating the final mask for the target. 
Such an approach suffers from two limitations: i) the need of accurate annotations; ii) the sequential/independent nature of the mapping between the atlases and the target image. 
\\In this work we propose a \coregistration{} and \cosegmentation{} framework that optimally aligns and segments a \emph{set} of input volumes. 
We adopt the standard graph-based deformable registration framework of Glocker et al.~\cite{glo11}. 
Our novel energy formulation incorporates discriminative information produced by alternative classifiers, trained to differentiate between different cortical structures. 
We stress the fact that our method is different than typical MAS: the final segmentations are obtained after population registration, while the probabilistic segmentations delivered by our classifiers are used as a discriminative image representation that helps to improve registration performance.
Therefore our approach is able to deal with the bias introduced from inaccurate segmentations while at the same time it exploits knowledge of the entire dataset simultaneously. Previous works on groupwise registration and segmentation of MR images have relied on image similarites~\cite{bhatia2004consistent}, or shape and texture models~\cite{tsai2004mutual,babalola2006groupwise}.
The works that are most similar to ours are~\cite{Heckemann2010},~\cite{Alchatzidis2014Discrete} and~\cite{Parisot2012}. 
They use probabilistic priors obtained with a pre-trained classifier to improve segmentation and registration. However, rather than performing prior- and intensity-based registration steps independently, as in~\cite{Heckemann2010}, we consider both types of data at the same time in a single, compound matching criterion. 
Furthermore, in~\cite{Alchatzidis2014Discrete} and~\cite{Parisot2012} segmentation variables are explicitly modeled, whereas we only model registration variables, thus reducing the number of parameters to be estimated. 
More importantly, these works aim at segmenting a single target image; contrary to that, we consider a target \emph{population} of images to be segmented and registered simultaneously. \\
Our method infers the segmentations of the unseen images on-the-fly using learned classifiers, and incorporates this information in the energy formulation. As our experimental results in Section 3.1 demonstrate, our method has considerable advantages over standard MAS as well. Firstly, given a set of target volumes, MAS would repeatedly register a set of ground truth masks and perform label fusion individually for each target. Contrary to that, we compute the segmentation probabilities once, and then segment all the volumes simultaneously. If numerous ground truth masks are to be used for the registration step, our method leads to substantial computational gains, as complexity depends only on the number of volumes we want to segment. Secondly, in the case of large datasets, the burden of selecting an appropriate ground truth subset to perform MAS more efficiently is removed; one simply has to compute the probability masks on the input volumes. Thirdly, in typical MAS only appearance features are used to compute the deformation fields between source and target. We go one step further, exploiting more sophisticated, learned representations to drive the coregistration process. These features are computed for all volumes involved, and are directly related to the desired final output. We validate the effectiveness of our approach on the task of segmenting challenging sub-cortical structures in two different brain imaging datasets.

\section{Problem formulation using segmentation priors}\label{sec:method}
We formulate our \coregistration\ and \cosegmentation\ algorithm as an energy minimization problem. 
The input is a set of 3D images $\imgs=\{ I_1, I_2,\ldots, I_N \},\ \ I_i:\Omega\subset\mathbb{R}^{3}\rightarrow\mathbb{R}$, and their corresponding segmentation likelihoods $\segs=\{ S_1, S_2,\ldots, S_N \}$ associated to the possible segmentation classes $c \in \mathcal{C} = \{ 0, \ldots, C \}$ as $S_i:\Omega \times \mathcal{C} \rightarrow[0,1]$. Label zero (0) corresponds to the background.
The output is the final multi-label segmentation masks $\hat{\segs}=\left \{ \hat{S}_1, \hat{S}_2, {...}, \hat{S}_N \right \}$ together with the deformation fields $\hat{\defs}=\left \{ \hat{T}_1, \hat{T}_2, {...}, \hat{T}_N \right \}$ that warp every image to a common coordinate space through an operation $I\circ \hat T$. 
In addition, let $\delta_\mathcal{X}$ be a function that measures similarity between inputs that lie in some domain $\mathcal{X}$.  
The objective function we want to minimize is 
\begin{eqnarray}
  {E}(\defs; \imgs, \segs) =  E_I\left( \defs; \imgs \right) + E_S\left (\defs; \segs\right) + E_R\left(\defs\right). 
  \label{eq:energyCoRegSeg}
\end{eqnarray}
The first two terms seek agreement on the appearance of equivalent voxels and deformed priors respectively, across all volumes of the registered population:
\begin{eqnarray}
  E_I(\defs; \imgs) & = & \sum_{x \in \Omega} \delta_\imgs ( I_1 \circ T_1(x), I_2 \circ T_2(x), \dots, I_N \circ T_N(x)), \\
  E_S(\defs; \segs) & = & \sum_{c \in \mathcal{C}} \sum_{x \in \Omega} \delta_\segs( S_1 \circ T_1(x, c), S_2 \circ T_2(x, c), \dots, S_N \circ T_N(x, c)).
  \label{eq:energyCoRegSegIntensities}
\end{eqnarray}
Here, $\delta_\imgs$ and  $\delta_\segs$ can be viewed as generalizations of the pairwise similarity, so as to account for multiple inputs. 
The deformation fields are applied on the probability map of each label separately and in the end we sum over all possible semantic labels $c \in \mathcal{C}$.
 
The last term, $E_R$, imposes geometric or anatomical constraints on the deformation fields, e.g. smoothness. 
Different types of regularizers $\mathcal{R}$ can be used, usually chosen as convex functions of the gradient of the deformation field. We describe our choice of $\delta_\imgs$, $\delta_\segs$ and $\mathcal{R}$ in~\refsec{sec:MRF}. We apply $\mathcal{R}$ to each deformation field $T_i$ independently: 
\begin{eqnarray}
E_R(\defs) =  \sum_{i=1}^N \sum_{x \in \Omega} \mathcal{R}( T_i(x) ). 
\label{eq:energyCoRegSegRegularizer}
\end{eqnarray}
By minimizing the energy defined in~\refeq{eq:energyCoRegSeg} with respect to $\defs$, we can obtain the optimal deformation fields $\hat{\defs}=\argmin_{\defs} E(\defs; \imgs,\segs).$ 
The high-order terms that appear in $E_I$ and $E_S$ are hard to optimize and diminish the guarantees to obtain the globally optimal solution. As a remedy we propose the two-step procedure adopted from~\cite{sot09}.
Instead of considering all the deformation fields at the same time, we estimate the deformation field $T_k$ of a single image, keeping all other images $(i \neq k)$ fixed.
This process is iterated for $i=1,2,\ldots,N$, and is reminiscent of the $\alpha$-expansion algorithm~\cite{boy01}: we start with an initial solution (in our case, the identity deformation fields) and iteratively move towards the optimal deformation fields that minimize $E$.

Once the optimal deformation fields $\hat{\defs}$ have been estimated, we can build the final segmentation masks $\hat{\segs}$. We first warp all segmentation priors in $\segs$ to the common frame of reference, generating the deformed segmentation masks $S_i \circ \hat T_i$. 
Then, given a target volume $I_k$ whose final segmentation we want to estimate, we back-project all warped segmentation masks $S_i \circ \hat T_i$ from the common frame, to the coordinate space of $I_k$ using the inverse deformation field $T_k^{-1}$.
This method is modular with respect to the fusion strategy. We use a simple majority voting, assigning to every voxel the class $c \in \{ 0,\ldots, C \}$ with the highest number of votes after back-projection.\\

\noindent \textbf{Iterative Algorithm. }\label{sec:chapterSemantic:coreg:discreteFormulation}
We now rewrite~\refeq{eq:energyCoRegSeg} as an iterative process. $E^{t}_I$, $E^{t}_S$ and $E^{t}_R$ consider a single deformation field $T_k$ at a time $t$ and are computed as 
\begin{eqnarray}
E^{t+1}_I (T^t_k; \imgs ) & = & \sum_{i=0,i\neq k}^{N}\sum_{x \in \Omega }\delta_\imgs ( I^t_i, I^t_k \circ T^t_k(x) ) \label{eq:M}\\
E^{t+1}_S(T^t_k; \segs)& = &\sum_{i=0,i\neq k}^{N} \sum_{c \in \mathcal{C}} \sum_{x \in \Omega }\delta_\segs ( S^t_i, S^t_k \circ T^t_k(x, c) ) \label{eq:energyCoRegSegPriorsDiscrete} \\
E^{t+1}_R(T^t_k) & = & \left ( N-1 \right )\sum_{x \in \Omega }\mathcal{R}(T^t_k(x)).
\end{eqnarray}
$I^t,S^t,T^t$, denote the current image, segmentation and deformation field respectively, after applying the updates at iterations $1,2,\ldots,t$.
The regularization term is scaled by $(N-1)$ for normalization purposes. 
This iterative process is repeated until convergence. After all images have been aligned in a common reference frame, majority voting produces the final segmentation masks.
For clarity, in the remaining of the text we drop the dependence on $t$.
A step-by-step description of the procedure is given in Algorithm~\ref{alg:ICSP}.

\begin{algorithm*}[t!]
\caption{Iterative Coregistration-Cosegmentation algorithm}
\label{alg:ICSP}
\fontsize{9}{11}\selectfont
\begin{algorithmic}[1]
\Procedure{ICS}{$\imgs = \left \{ I_1, I_2, \ldots , I_N \right \}$, $\segs =  \left \{ S_1, S_2, \ldots, S_N \right \}$}
		\State Initialize the deformation fields $\left \{ \hat{T}_1, \hat{T}_2, {...}, \hat{T}_N \right \}$ as null (identity) deformation fields
		\Repeat
		\Repeat
		\State Sample an image $I_k \in \imgs$ without replacement
		\State Register $I_k$ to all images in $\imgs \setminus \{I_k\}$, optimizing $E$: 
		\begin{eqnarray}
		\ddot{T}_k = \argmin_{T_k}  E_I(T_k; \imgs) + E_S( T_k;\segs) + E_R( T_k )
		\label{eq:singleRegistrationProblem}
		\end{eqnarray}
		\State Deform image and corresponding segmentation: $I_k \leftarrow I_k \circ \ddot{T}_k,\ S_k \leftarrow S_k \circ \ddot{T}_k$
		\State Update deformation field $\hat{T}_k \leftarrow \hat{T}_k \circ \ddot{T}_k$
		\Until{all images have been chosen once} 
		\Until{All $T$ remain unchanged or the maximum of iterations is reached}
		\For{\text{each image $I_k\in \imgs$}}
		\For{\text{each segmentation prior $ S_i \in \segs $}} 
  		\State \text{Deform ${S}_i$ to the native space of ${I}_k$: $ S'_i = {S}_{i} \circ \hat{T}_{k}^{-1}$}		
      		\EndFor
		\State Apply label fusion (e.g., \textit{Majority Voting}) on $\{ S'_i\}_{i \in \{ 1, \ldots, N \}}$ to obtain $\hat S_k$
		\EndFor
		\State Output: $\hat{\defs}=\left \{ \hat{T}_1, \hat{T}_2, {...}, \hat{T}_N \right \}$ and $\hat\segs = \left \{ \hat{S}_1, \hat{S}_2, {...}, \hat{S}_N \right \}$ 
\EndProcedure
    \end{algorithmic}
\end{algorithm*}

\noindent \textbf{Discrete Formulation. }\label{sec:MRF}
We formulate non-rigid registration between two images $I_i, I_k$ as a discrete energy minimization problem. 
Following~\cite{Rueckert1999Nonrigid}, we parametrize the deformation fields $T_k$ as a linear combination of $K \ll |\Omega|$ control points that form a regular 3D grid.
We define a first order discrete MRF by superimposing an undirected graph $G = (V, U)$ on an image, with $V$ and $U$ denoting the graph nodes and edges respectively. 
Nodes are interpreted as random variables that model displacements $\mathbf{d}_p \in \mathbb{R}^3$ of the control points, while edges encode the interaction between these variables, in a 6-way neighborhood $U_p$.

Given a labeling $L = \{ l_1, l_2, \ldots, l_K \} = \{\mathbf{d}_1, \mathbf{d}_2, \ldots, \mathbf{d}_K \}$, that assigns a label (displacement vector) to every node $p$ in the MRF, the energy function becomes
\begin{eqnarray}
E_{\mathrm{MRF}}(L;G) & = & \sum_{p\in V}{g_p(l_p) }+\lambda \sum_{(p,q)\in U_p}{f_{pq}(l_p, l_q)},\ \ \text{where}\\
g_p(l_p) = g_p(\mathbf{d}_p)  & = &\sum_{x \in \Omega_p} \hspace{-1mm}\delta_\imgs ( I_i, I_k \circ T_{k}^{\mathbf{d}_p} (x))  +  \beta \sum_{c \in \mathcal{C}}  \sum_{x \in \Omega_p}\hspace{-1mm} \delta_\segs (S_i, S_k \circ T_{k}^{\mathbf{d}_p} ( x, c ) ).\nonumber
\label{eq:EMRF}
\end{eqnarray}
The unary term $g_p$ is a combination of terms $E_I,E_S$ that encode appearance and segmentation likelihood agreement.
In practice, control points have a limited spatial support, therefore $p$ receives contributions only from pixels inside a region $\Omega_p$ (e.g. patch) around it.
$T_{k}^{\mathbf{d}_p}$ is the transformation induced by applying the displacement vector $\mathbf{d}_p$ on the control point $p$. 
The $\beta$ coefficient determines the influence of segmentation priors on the optimization problem and $\lambda$ is a scaling factor. In our experiments we set $\lambda=5, \beta=100$ using cross-validation. As $\delta_\imgs$ we use the sum of absolute difference (SAD), while $\delta_\segs$ computes the Hamming distance on the segmentation maps obtained after assigning the semantic class with highest probability to each pixel.
The pairwise term $f_{pq}( l_p,l_q ) = f_{pq}( \mathbf{d}_p,\mathbf{d}_q ) = || \mathbf{d}_p-\mathbf{d}_q ||$ is a discrete approximation of the gradient of the spatial transformation and acts as the regularizer $\mathcal{R}$ in~\refeq{eq:energyCoRegSegRegularizer}. 

To infer the best labeling, we employ Fast-PD~\cite{Komodakis2008Performance}, an efficient move-making discrete optimization method based on linear programming relaxation, that has shown promising results when applied to multi-label problems with similar types of energies.

\section{Experiments}\label{sec:experiments}
We evaluate the performance of our approach on the task of subcortical brain structure segmentation on two MRI datasets, IBSR~\cite{Rohlfing2012Image} (18 subjects, slice tickness of 1.3mm) and a Rolandic Epilepsy (RE) study (35 subjects, slice tickness of 1mm).
In our experiments we use two types of classifiers to estimate segmentation maps, which are then used to guide the registration: convolutional neural networks (CNNs) and random forests (RFs). 
For a description on the CNN architecture, training methodology and RE dataset, we refer to~\cite{Shakeri2016}. 
We focus on a subset of $16$ subcortical structures, including left and right lateral ventricle, thalamus, caudate, putamen, pallidum, hippocampus, amygdala, and accumbens. Below we list the variants compared in our experiments.

\begin{figure*}[t!]
\centering
\includegraphics[width=\textwidth]{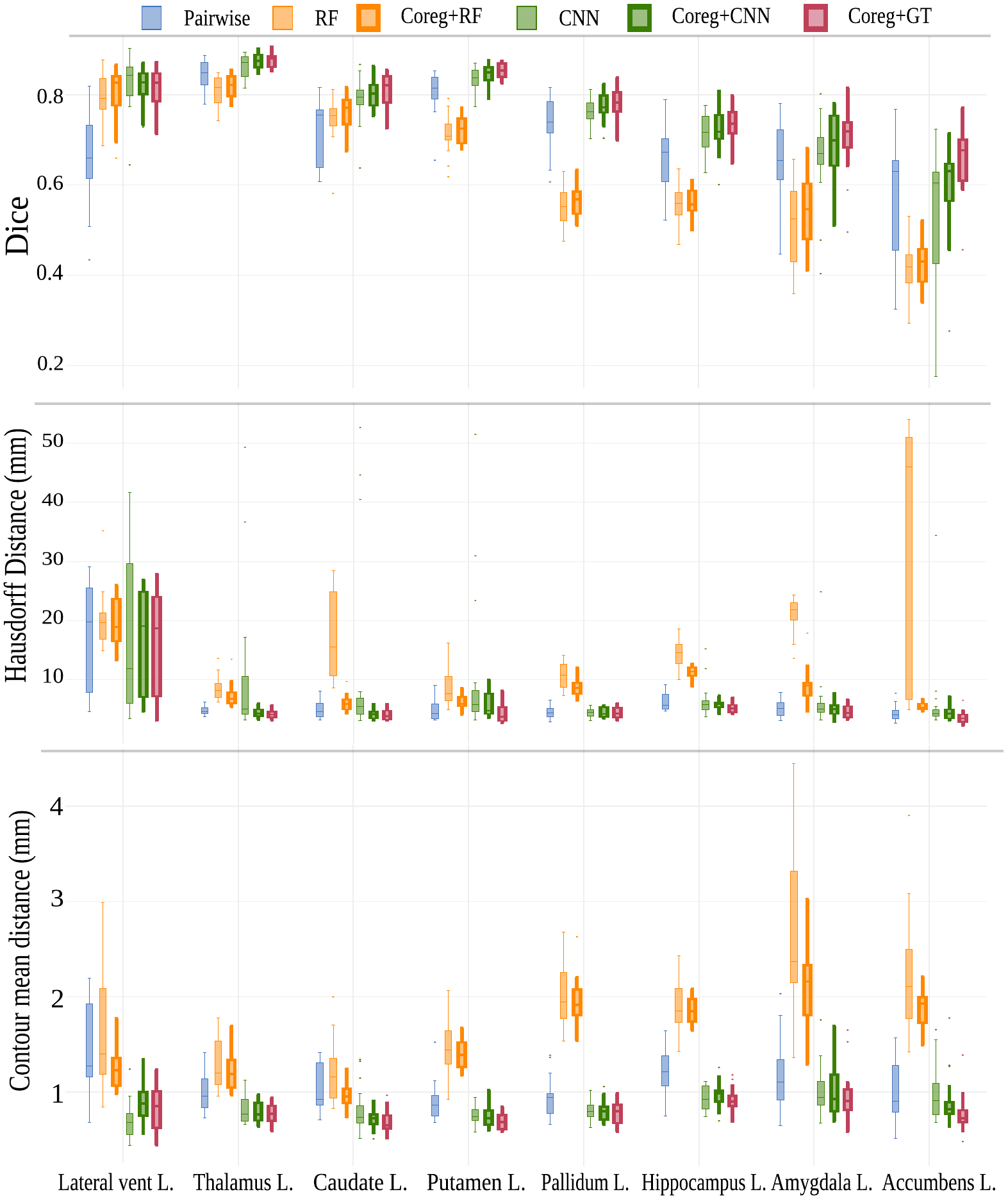}
\caption{Box plots for average Dice coefficient (DC), Hausdorff distance (HD) and contour mean distance (CMD) for left side subcortical structures in IBSR (best viewed in color). \textbf{DC:} higher = better. \textbf{HD/CMD:} lower = better. Results for the right-side are included in the supplementary.}
\label{fig:resultsIBSR}
\end{figure*}

\noindent \textbf{Coreg+CNN and Coreg+RF:} We use the terms Coreg+CNN and Coreg+RF to refer to the variants of our method that use CNN and RF priors respectively. 
We generate the CNN and RF priors using the methods described in~\cite{Shakeri2016} and~\cite{Alchatzidis2014Discrete} respectively.

\noindent \textbf{CNN and RF:} To further demonstrate the effect of using the iterative \coregistration{} on top of CNN/RF priors, we report segmentation results based on the CNN/RF probability maps without \coregistration{}. In this setting, given a CNN/RF prior, the segmentation class of every voxel is simply chosen as the class with the highest probability. 

\begin{figure*}[t!]
\centering
\includegraphics[width=\textwidth]{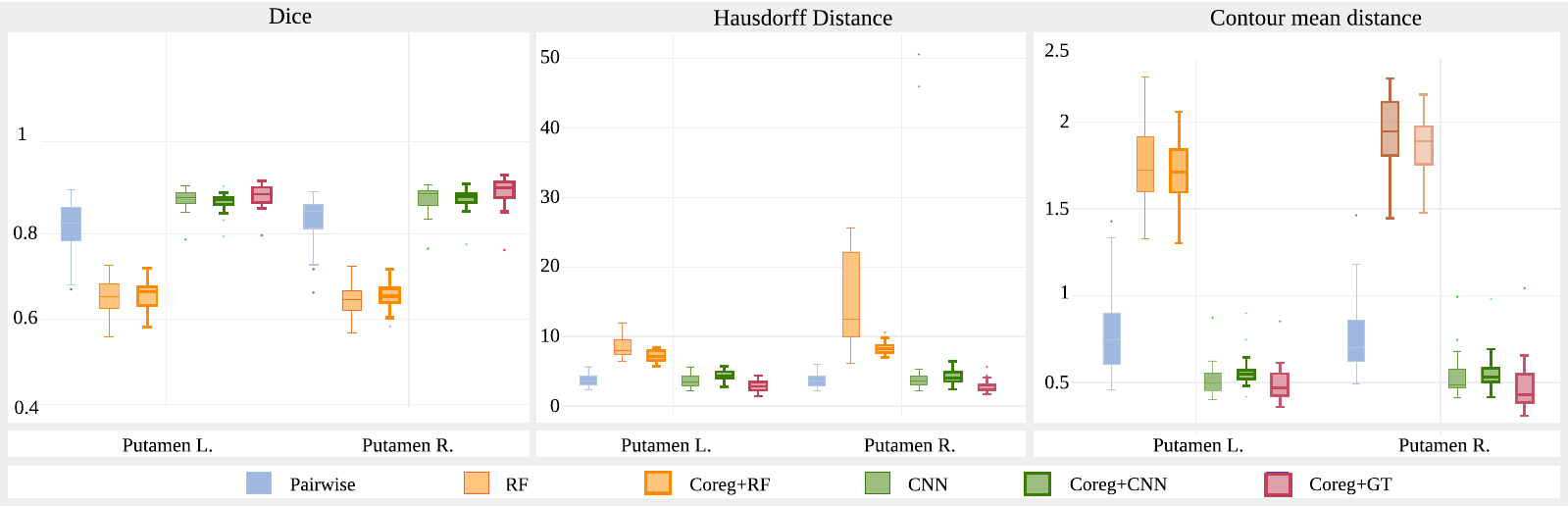}
\caption{Box plots for average Dice coefficient (DC), Hausdorff distance (HD), and contour mean distance (CMD)for the left and right putamen in the RE dataset (best viewed in color). \textbf{DC:} higher = better. \textbf{HD/CMD:} lower = better. Coreg+CNN results approach the performance of the oracle.}
\label{fig:MTL}
\end{figure*}

\noindent \textbf{Pairwise:} As a baseline, we implement the standard MAS based on pairwise registration. All atlases are independently registered to the target image as in~\cite{glo11}; then the \groundtruth\ annotations are fused to generate the final segmentation using majority voting.
The use of the actual \groundtruth\ annotations offers a clear advantage with respect to Coreg+CNN and Coreg+RF, that use the \emph{estimated} segmentation probability maps instead. Still, Coreg+CNN achieves better performance as shown in in Figures~\ref{fig:resultsIBSR},\ref{fig:MTL}.

\noindent \textbf{Coreg+GT (Oracle):} 
The merit of our approach is that it allows us to guide the \coregistration\ process using probability maps as a surrogate for \groundtruth\ annotations, which are not always available. 
In order to assess the maximum potential of our method, we implemented an \emph{oracle} that provides us with an upper-bound to its performance. 
The oracle makes use of the \groundtruth\ segmentation masks for all 3D volumes, except for the target image, for which we keep the probability maps computed by the CNN. \\

\begin{figure}[t!]
\centering
\includegraphics[scale=0.29]{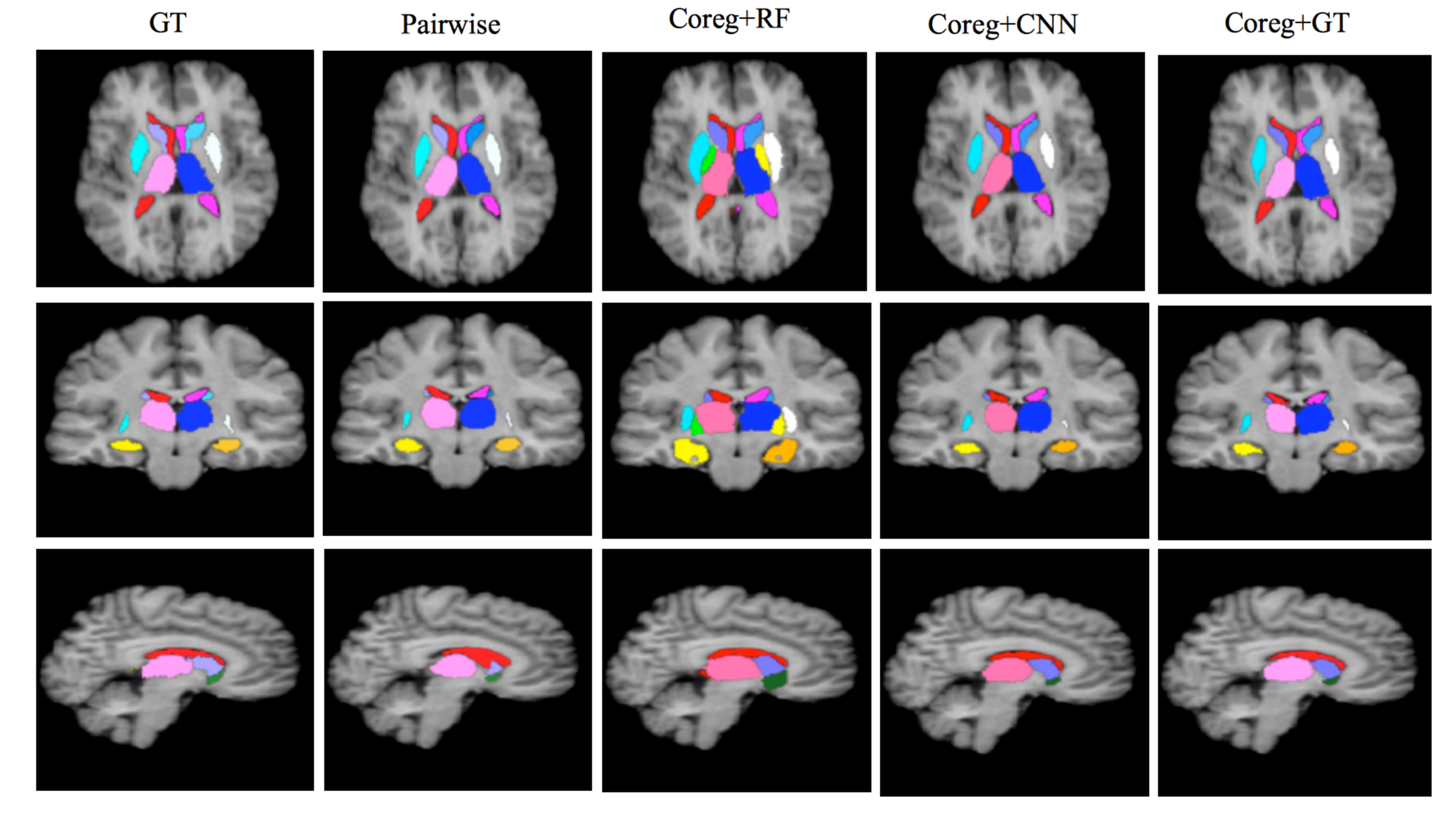}
\vspace{-5mm}
\caption{Segmentation results in three different views. 
Coreg+CNN can be used as a reliable substitute for \groundtruth\ annotations in multi-atlas \coregistration{} and \cosegmentation{} (view in color).}
\label{fig:visualResult}
\end{figure}

We summarize the results of our experiments in Figures~\ref{fig:resultsIBSR}-\ref{fig:MTL}.
We compare performance using three different metrics: i) average Dice coefficient (DC); ii)Hausdorff distance (HD); iii) contour mean distance (CMD).
Our results show that Coreg+CNN achieves higher segmentation accuracy compared to both Coreg+RF and the pairwise segmentation baseline. Respectively, the segmentations obtained using only the CNN classifier output (without any registration process) are much more accurate than the ones from random forests. Unsurprisingly, Coreg+GT outperforms all other variants. Nonetheless, performance of Coreg+CNN is close to Coreg+GT in most cases, also illustrated visually in~\reffig{fig:visualResult}. 
This evidence solidifies our original claim, that reliable priors can act as a practical substitute for golden standard annotations in multi-atlas segmentation. 

Another important observation is that our \coregistration{} and \cosegmentation{} framework significantly improves results of less accurate priors (e.g. the ones produced by RF), especially in terms of Hausdorff and contour mean distance.  
Such priors can be learned from weak annotations that are produced very efficiently compared to precise segmentation masks (e.g. bounding boxes) and still deliver acceptable results.

\section{Conclusions}\label{sec:conclusions}
In this paper we have proposed a novel method for \cosegmentation\ and \coregistration\ of multi-volume data, guided by semantic label likelihoods. Our approach has the following characteristics: i) infers deformations that are anatomically plausible; ii) establishes visual consistencies between all volumes according to any metric; iii) enforces segmentation consistencies among all volumes according to the predicted likelihoods. 
Experimental evaluation on a standard, publicly available benchmark, as well as on an additional clinical dataset, demonstrates the effectiveness of our approach. Our experiments also show the value of reliable segmentation priors. Label likelihoods extracted with a deep CNN outperform alternative methods and can replace \groundtruth{} annotations in \coregistration{} with minimal loss in performance. 

Future research directions include studying the gains of combining different metrics per class and using them as content-adaptive potentials in the energy function. 
Explicitly modeling high-order interactions and simultaneously recovering all deformations with one-shot optimization are also of great theoretical and practical interest. 
Finally, an important future goal is testing the proposed method on a clinical problem where \coregistration\ and \cosegmentation\ are important, such as adaptive radiotherapy.

\bibliographystyle{splncs}
\bibliography{registrationBib2}

\end{document}